\pdfoutput=1

\documentclass[11pt]{article}

\usepackage[]{acl}

\usepackage{times}
\usepackage{latexsym}

\usepackage[T1]{fontenc}

\usepackage[utf8]{inputenc}

\usepackage{microtype}

\usepackage{inconsolata}
\usepackage{graphicx}
\usepackage{url}
\usepackage{enumitem}
\usepackage{booktabs}
\usepackage{multirow,stackengine}
\usepackage{bbding}
\usepackage{soul}
\definecolor{llightgray}{RGB}{230,230,230}
\title{SlideAVSR: A Dataset of Paper Explanation Videos\\for Audio-Visual Speech Recognition}
\author{Hao Wang$^{1}$\; Shuhei Kurita$^2$\; Shuichiro Shimizu$^3$\; Daisuke Kawahara$^{1,2,4}$\\
$^1$ Waseda University\; $^2$ National Institute of Informatics (NII)\; $^3$ Kyoto University\; $^4$ LLMC, NII\\
\texttt{conan1024hao@akane.waseda.jp}
\;\texttt{skurita@nii.ac.jp}\\
\texttt{sshimizu@nlp.ist.i.kyoto-u.ac.jp}
\;\texttt{dkw@waseda.jp}\\
}
\begin{document}
\maketitle
\begin{abstract}


Audio-visual speech recognition (AVSR) is a multimodal extension of automatic speech recognition (ASR), using video as a complement to audio.
In AVSR, considerable efforts have been directed at datasets for facial features such as lip-readings, while they often fall short in evaluating the image comprehension capabilities in broader contexts.
In this paper, we construct \textbf{SlideAVSR}, an AVSR dataset using scientific paper explanation videos.
SlideAVSR provides a new benchmark where models transcribe speech utterances with texts on the slides on the presentation recordings.
As technical terminologies that are frequent in paper explanations are notoriously challenging to transcribe without reference texts, our SlideAVSR dataset spotlights a new aspect of AVSR problems.
As a simple yet effective baseline, we propose DocWhisper, an AVSR model that can refer to textual information from slides, and confirm its effectiveness on SlideAVSR.
\end{abstract}
\vspace{-10pt}
\section{Introduction}
\vspace{-5pt}
Research on multimodal models capable of handling multiple types of data, such as language, images, videos, and audio simultaneously, has garnered significant attention.
An example is audio-visual speech recognition (AVSR), a multimodal extension of automatic speech recognition (ASR), using video as a complement to audio.
Most previous studies in AVSR have been conducted with the aim of improving accuracy on lip reading datasets~\citep{8585066,afouras2018lrs3ted}.
While models built in these studies~\citep{shi2022avhubert,pan-etal-2022-leveraging,haliassos2023jointly} demonstrate high performance on lip reading data, their applicability to other types of videos remains limited.

In this paper, we aim to evaluate the image comprehension capabilities of AVSR models across a broader spectrum of visual contents than facial features.
To achieve this, we construct SlideAVSR, an AVSR dataset that contains various technical terms that are notoriously challenging to transcribe without referring to textual information on slides.
Specifically, we collect scientific paper explanation videos from YouTube, apply data refinement procedures with several custom filters, and perform data partitioning considering the speakers' accents. 

Furthermore, we propose DocWhisper, a simple yet effective AVSR baseline that can efficiently refer to the content of slides using optical character recognition (OCR).
In experiments utilizing SlideAVSR, DocWhisper demonstrated a performance improvement of up to 14.3\% compared to Whisper~\citep{radford2022robust}, which relies solely on audio input.
Additionally, to address the long-tail problem in OCR results, we introduce FQ Ranker, which calculates word ranks based on the frequency of word occurrences, and we evaluate its effectiveness integrated with DocWhisper.
\vspace{-2pt}
\section{Related Work}
\vspace{-2pt}
Compared to the efforts that have been made on lip reading datasets~\citep{8099850,inproceedingsChung,10.1007/978-3-319-54184-6_6,8585066,afouras2018lrs3ted,shillingford19_interspeech}, AVSR datasets in other types of videos remain scarce.
VisSpeech~\citep{gabeur2022avatar} is constructed from a subset of the instructional video dataset HowTo100M~\citep{miech19howto100m} where the visual stream and speech audio are semantically related.
The audiovisual diarization benchmark in the Ego4D challenge~\citep{Ego4D} consists of 585 egocentric video clips.
In terms of utilizing textual information extracted from videos, SlideSpeech~\cite{wang2023slidespeech} builds an AVSR dataset on online conference videos enriched with slides.
However, the filtering process of SlideSpeech relies heavily on human annotators.
In this work, we exclude videos and utterances that do not correspond to slides by utilizing a vision language model.
This approach reduces annotation costs and enhances the purity and quality of our dataset within the slide domain.

In the context of extending Whisper to an AVSR model, \citet{peng2023whisper} employed CLIP~\citep{radford2021learning} to transform the input visual stream into word sequences, which were then utilized as prompts for Whisper.
They reported that this approach enhances the zero-shot performance on VisSpeech.
In this study, we employ OCR to create prompts and implement fine-tuning to improve performance rather than using zero-shot prompting.
\section{SlideAVSR: Dataset Construction}
\begin{figure}[t]
    \centering
    \includegraphics[width=1\linewidth]{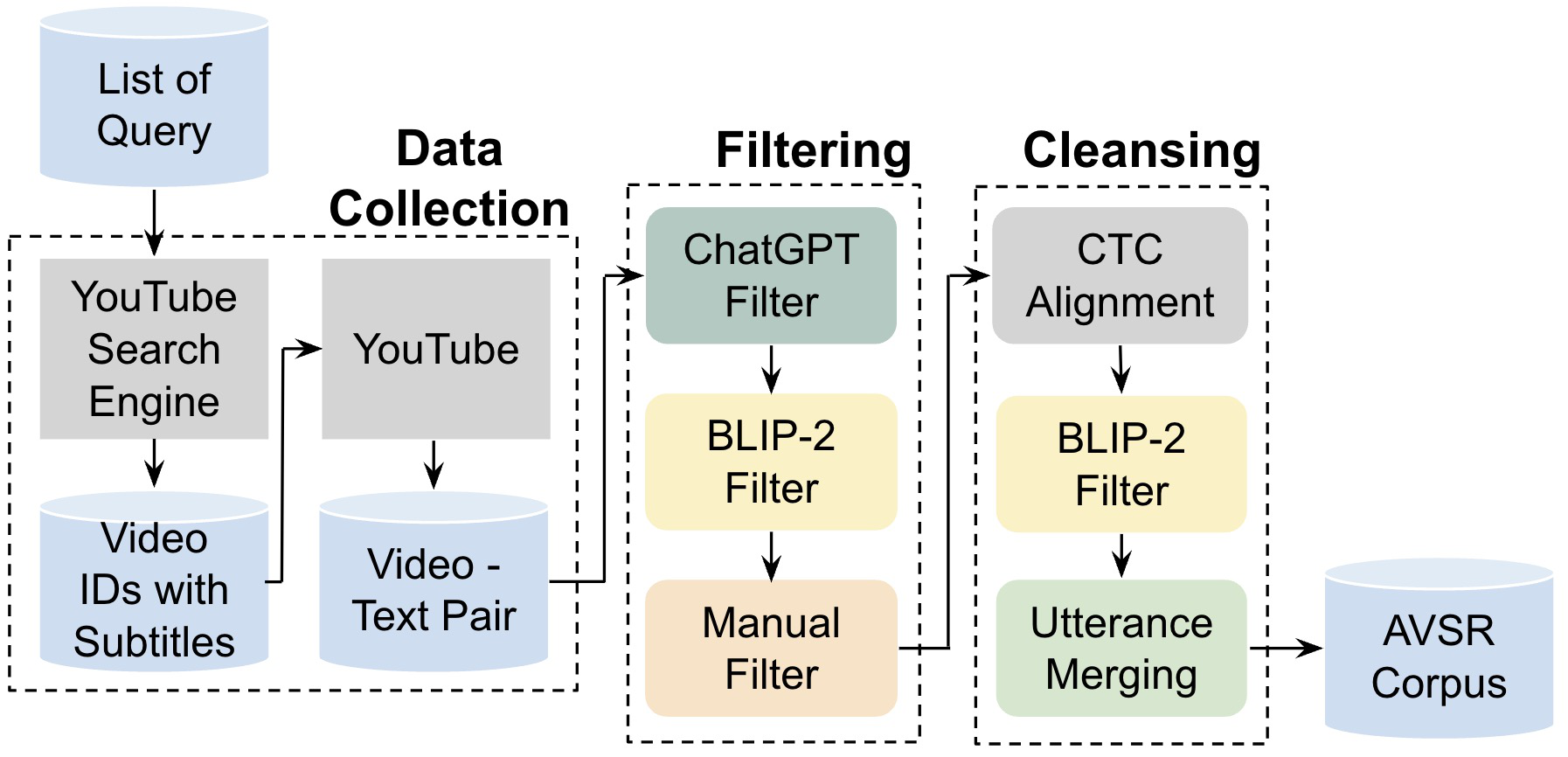}
    \caption{Construction flow of SlideAVSR.}
    \label{fig:pipeline}
    \vspace{-15pt}
\end{figure}

In this study, we construct SlideAVSR, an AVSR dataset based on scientific paper explanation videos incorporating various technical terms, making accurate transcription difficult without referring to the slides.
Based on JTubeSpeech~\cite{takamichi2021jtubespeech}, a framework for building audio corpora from YouTube videos, we implement several custom filters to target videos, thereby applying high-precision data refinement.
This section describes the construction flow of SlideAVSR. 
Figure~\ref{fig:pipeline} illustrates the flow.
\subsection{Data Collection}
\paragraph{Creating search queries.}
We first collect videos with search queries that are related to top conferences in the field of artificial intelligence.
We create queries in the format \colorbox{llightgray}{\{\hspace{-1pt}Conference\hspace{-1pt}\}}
\colorbox{llightgray}{\{\hspace{-1pt}Year\hspace{-1pt}\} \{\hspace{-1pt}Form\hspace{-1pt}\}}.
The list of target conferences is provided in Appendix~\ref{appendix:conf_list}.
Considering the increased prevalence of online conferences since COVID-19, we focus on the years 2020 to 2023.
The forms include ``paper'', ``workshop'', and ``talk''.
An example search query is ``ACL 2023 paper''.
\paragraph{Obtaining videos with subtitles.}
Using the search queries, we retrieve video IDs with subtitles and download them.\footnote{https://github.com/yt-dlp/yt-dlp}
To ensure data quality, only videos with manual subtitles are considered.
Additionally, we set the following criteria:
\begin{itemize}[topsep=5pt,after=\vspace{-3pt}]
    \setlength{\parskip}{0cm}
    \setlength{\itemsep}{0cm}
    \item Duration between 5 and 20 minutes (videos that are too short or too long are less likely to be paper explanation videos).
    \item Video format: MP4, 720P, H264.
    \item Audio format: single-channel, 16bit, 16kHz.
\end{itemize}
A total of 636 videos were downloaded.
\subsection{Filtering}
\label{sec:filtering}
We curate several filters to remove videos that are not paper explanations or do not include slides.
\paragraph{ChatGPT filter.}
We provide the videos' description for ChatGPT\footnote{https://openai.com/product} to confirm the following:
\begin{itemize}[topsep=5pt,after=\vspace{-3pt}]
    \setlength{\parskip}{0cm}
    \setlength{\itemsep}{0cm}
    \item This video is an explanation of a paper.
    \item The description is written in English.
\end{itemize}
We perform three times of generation, and if ``Yes'' is outputted at least once, we adopt the video; otherwise, we discard it.
We show the details of the model and prompt in Appendix~\ref{appendix:data_model_and_prompt}.
A total of 342 videos were excluded, leaving 294 videos for subsequent processes.

\paragraph{BLIP-2 filter for videos.}
We capture screenshots at the beginning, end, and three quartile points in the timeline for each video, and then present these screenshots to the vision language model BLIP-2~\cite{li2023blip2} to verify the following:
\begin{itemize}[topsep=5pt,after=\vspace{-3pt}]
    \setlength{\parskip}{0cm}
    \setlength{\itemsep}{0cm}
    \item This image is a screenshot, not a photo.
    \item This image is a part of slides.
\end{itemize}
We perform generation for each screenshot, and if ``Yes'' is outputted at least once, we adopt the video; otherwise, we discard it.
We show the details of the model and prompt in Appendix~\ref{appendix:data_model_and_prompt}.
A total of 6 videos were excluded, leaving 288 videos for subsequent processes.

\paragraph{Manual filter.}
We conduct manual checks to remove inappropriate videos that are not excluded by the automatic filters, including:
\begin{itemize}[topsep=5pt,after=\vspace{-3pt}]
    \setlength{\parskip}{0cm}
    \setlength{\itemsep}{0cm}
    \item Videos rarely showing slides.
    \item Videos unrelated to paper explanations, such as conference openings.
\end{itemize}
A total of 38 videos were excluded, leaving 250 videos for subsequent processes.
\subsection{Cleansing}
\label{sec:cleansing}
We implement audio-subtitle alignment, exclude utterances that do not correspond to slides, and merge short utterances for data cleansing.
\paragraph{CTC alignment.}
Due to the inaccuracy in the timing of subtitles, we implement audio-subtitle alignment and scoring using CTC segmentation~\cite{kurzinger2020ctc}. 
We set the threshold to -7 and exclude utterances with lower scores.
The details of the model are shown in Appendix~\ref{appendix:data_model_and_prompt}.
Approximately 2.5\% of utterances were excluded through this process.

\paragraph{BLIP-2 filter for utterances.}
We capture screenshots at the midpoint of each utterance, followed by filtering using BLIP-2.
Three generations are conducted for each screenshot, and if ``Yes'' is outputted at least once, we adopt the utterance; otherwise, we discard it.
The employed prompt is identical to the BLIP-2 filter in Section~\ref{sec:filtering}.
Approximately 1.0\% of utterances were excluded through this process.

\paragraph{Merging utterances.}
Subtitles created by video authors occasionally exhibit unnatural segmentation, resulting in exceedingly brief spans.
Utilizing the audio segments obtained through CTC segmentation, we implement a merging process, combining two consecutive utterances into a single entity if the end time of the preceding utterance aligns with the start time of the subsequent one and their cumulative duration does not exceed 15 seconds.
This procedure significantly enhanced Whisper's ASR performance by approximately 20\%.
\subsection{Data Partitioning}
\vspace{-2pt}
\label{sec:data partition}
Previous studies~\citep{meyer-etal-2020-artie,javed2023svarah,dichristofano2023global} have suggested that the performance of ASR systems significantly varies depending on the speaker's accent\footnote{The term ``accent'' in this paper refers to comprehensive prosodic information, including accent, intonation, tone, etc.}.
Based on the hypothesis that visual information contributes to the recognition of challenging accents, we ask native English speakers to classify the speakers' accents in SlideAVSR and perform dataset partitioning.
We partition the dataset into Train, Dev, and TestA, reserving a smaller yet significant TestB subset for South Asian English (SAE) accents.
During partitioning, we have ensured that the same speaker did not belong to multiple partitions.
Additionally, 5 videos with machine-generated audio were manually excluded by the annotators.

Through the construction flow, we produced an AVSR dataset of around 36 hours from 245 videos.
We show the statistics of the dataset in Table~\ref{tab:statistics}.

\begin{table}[t]
\centering
\resizebox{0.9\columnwidth}{!}{%
\begin{tabular}{lrrrr}
\hline
 & \#videos & \#speakers & \#utterances & \#hours \\
\hline
Train & 195 & 172 & 15,803 & 29.26 \\
Dev & 20 & 20 & 1,515 & 3.08 \\
TestA & 15 & 15 & 1,034 & 2.21 \\
TestB & 15 & 13 & 1,111 & 1.90 \\
\hline
Total & 245 & 220 & 19,463 & 36.45 \\
\hline
\end{tabular}}
\caption{Statistics of SlideAVSR.}
\label{tab:statistics}
\vspace{-15pt}
\end{table}
\section{Experiments}
\vspace{-2pt}
\subsection{Approaches}
DocWhisper processes the input video stream through an OCR module, extracting textual information into word sequences, which are then provided to Whisper as prompts for fine-tuning and inference.
While \citet{peng2023whisper} employed prompts derived from CLIP in zero-shot learning, our preliminary experiments did not reveal a performance improvement in zero-shot learning on SlideAVSR. 
Given that Whisper's pre-training~\citep{radford2022robust} did not use prompts, we speculate that Whisper loses robustness when it faces diverse prompts.

\begin{figure}[t]
    \centering
    \includegraphics[width=1\linewidth]{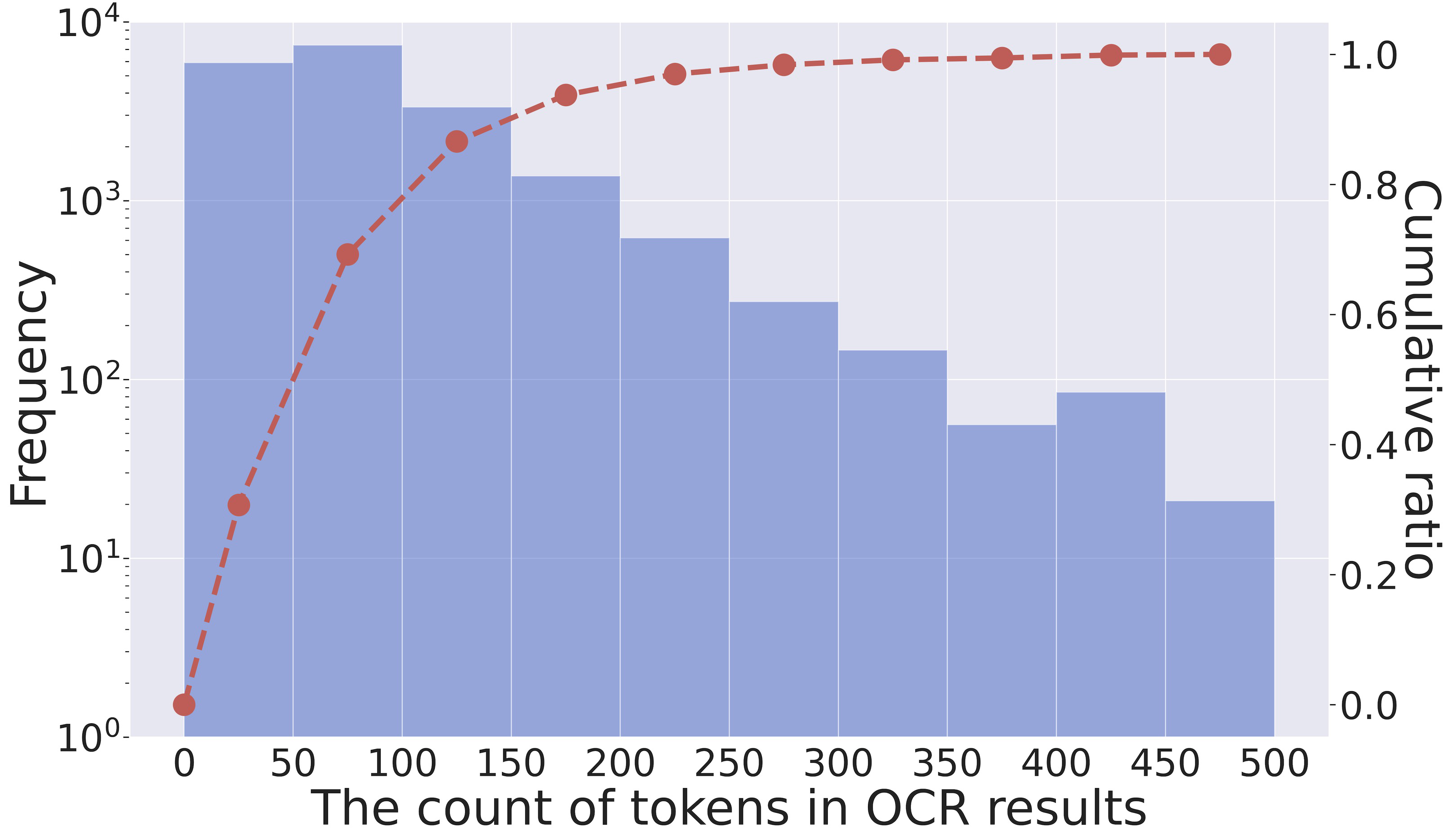}
    \caption{Frequency distribution of the number of words in OCR results. While samples with over 500 words are present, they are omitted for brevity.}
    \label{fig:count ocr}
    \vspace{-15pt}
\end{figure}

We show the frequency distribution of the number of words in OCR results in Figure~\ref{fig:count ocr}.
The distribution is long-tail, which means that only 70\% of the samples can be covered even if we include 100 words in the prompts\footnote{Whisper typically assigns a maximum length of 224 to prompts, making inputs with over 100 words challenging.}.
To address this issue, we propose FQ Ranker, which calculates word ranks based on the frequency of word occurrences.
Given the demonstrated high correlation between word frequency and familiarity as shown in previous studies~\citep{coltheart1981mrc,tanaka}, increasing the rank of less frequent and more challenging words is expected to enhance the information content of prompts.

\begin{table*}[t]
    \centering
    \resizebox{1.9\columnwidth}{!}{%
    \begin{tabular}{@{}l@{\quad}l@{}}
    \hline
        \multicolumn{1}{c}{Type} & \multicolumn{1}{c}{Example} \\\hline
        Technical term & \texttt{W Hyp: we select quantum \st{adhering} 2 and nxt as representative of pos protocols}\\
         (41\%) & \texttt{D Hyp: we select quantum {\color{magenta}{ethereum}} 2 and nxt as representative of pos protocols} \\        
        Inflection & \texttt{W Hyp: manual \st{transcript}\quad we call this setting supervised things we have paired data}\\
         (28\%) & \texttt{D Hyp: manual {\color{magenta}{transcripts}} we call this setting supervised things we have paired data}\\
        Mishearing & \texttt{W Hyp: we can also perform other tasks like \st{normal} view synthesis}\\
         (24\%) & \texttt{D Hyp: we can also perform other tasks like {\color{magenta}{novel}}\quad view synthesis}\\
        Name    & \texttt{W Hyp: this is a work done at ibm research with \st{gilmoseci chileo} and irina \st{rich}}\\
         (7\%) & \texttt{D Hyp: this is a work done at ibm research with {\color{magenta}{guillermo cecchi}} and irina {\color{magenta}{rish}}}\\
    \hline
    \end{tabular}}
    \vspace{-5pt}
    \caption{Error types and examples that are substitution errors in Whisper (W) but correct in DocWhisper (D).}
    \label{tab:error analysis}
    \vspace{-12pt}
\end{table*}

\begin{table}[t]
\centering
\resizebox{1\columnwidth}{!}{%
\begin{tabular}{l|ccr@{\hskip 0.2in}|@{\hskip 0.1in}r@{\hskip 0.2in}r}
\toprule
Model & Modality & Fine-tune & $K$$^a$ & TestA & TestB  \\
\midrule
\multirow{2}{*}{Whisper}    & \multirow{2}{*}{A}     & \XSolidBrush   &  \multirow{2}{*}{0}  & 8.23 & 11.18 \\
                            &                        & \CheckmarkBold &                      & 8.07 & 11.25 \\
\midrule
DocWhisper                 & \multirow{2}{*}{A $+$ V} & \multirow{2}{*}{\CheckmarkBold} & \multirow{2}{*}{25} &   \underline{7.35}&  10.82\\
\hspace{5pt} $+$ FQ Ranker &                          &                                 &                     &   7.42&  \underline{10.59}\\
\midrule
DocWhisper                 & \multirow{2}{*}{A $+$ V} & \multirow{2}{*}{\CheckmarkBold} & \multirow{2}{*}{50} &   \underline{7.08}&  10.43\\
\hspace{5pt} $+$ FQ Ranker &                          &                                 &                     &   7.26&  \underline{10.35}\\
\midrule
DocWhisper                 & \multirow{2}{*}{A $+$ V} & \multirow{2}{*}{\CheckmarkBold} & \multirow{2}{*}{75} &   \underline{7.02}&  \underline{10.04}\\
\hspace{5pt} $+$ FQ Ranker &                          &                                 &                     &   7.26&  10.29\\
\midrule
DocWhisper                 & \multirow{2}{*}{A $+$ V} & \multirow{2}{*}{\CheckmarkBold} & \multirow{2}{*}{100}&   \textbf{\underline{6.91}}&  \textbf{\underline{10.01}}\\
\hspace{5pt} $+$ FQ Ranker &                          &                                 &                     &   7.04&  10.22\\
\bottomrule
\end{tabular}}
\vspace{-5pt}
\raggedright
\footnotesize{$^a$}\footnotesize{Indicating maximum word counts for prompts.}\\
\caption{Quantitative evaluation (WER) on SlideAVSR.}
\label{tab:results}
\vspace{-15pt}
\end{table}
\subsection{Implement Details}
We used Whisper large-v3\footnote{https://huggingface.co/openai/whisper-large-v3} as a base model and Word Error Rate (WER) for evaluation.
In the case of DocWhisper, we captured screenshots at the midpoint of each utterance, fed them into the OCR module, and used the recognized text as the prompts to Whisper.
In this case, multiple utterances might share the same slide.
We use Google Cloud Vision API\footnote{https://cloud.google.com/vision} for OCR.
The prompts were presented to the model as word sequences, such as ``word 1, word 2, ..., word $n$''.
FQ Ranker utilized word frequency counts obtained from the English Wikipedia and sorted the OCR results in ascending order based on word frequency.
We conducted experiments with different maximum word counts for prompts ($K \in \{25, 50, 75, 100\}$) and with or without FQ Ranker.
More implementation details are provided in Appendix~\ref{appendix:implement details}.
\subsection{Results}
We show the results of quantitative evaluations for Whisper and DocWhisper in Table~\ref{tab:results}.
In both models, the scores of the TestB set, consisting of videos with SAE accents, were inferior to the scores of the TestA set, indicating that Whisper struggles with rare accents. 
With fine-tuning, Whisper demonstrated a 1.9\% improvement on the TestA set.
However, no notable improvement was observed for the TestB set.
Despite the presence of videos with SAE accents in the training data, their limited quantity was deemed insufficient to address the challenges posed by difficult accents.

Compared to the fine-tuned Whisper, DocWhisper exhibited a maximum improvement of 14.3\% on TestA and 11\% on TestB.
We gather that referring to textual information on slides can significantly improve speech recognition performance on SlideAVSR.
We also found that as the maximum word count of prompts increased, the performance improved, indicating that maximizing information content contributes to performance enhancement.

FQ Ranker improved the scores on TestB when the maximum word count of prompts was set to 25; however, this advantage was reversed when the maximum word count exceeded 50. 
Details provided in Section~\ref{sec:error analysis} indicate that transcriptions corrected by DocWhisper do not exclusively consist of technical terms, which suggests the potential for misinterpretation even in words with high familiarity.
We also speculate that sorting words based on word frequency disrupts the ordered contextual information, thus increasing the difficulty of Whisper's decoder, which is a language model, to refer to the textual information on the slides.
\subsection{Analysis of Specific Examples}
\label{sec:error analysis}
Among Whisper's errors (deletions, substitutions, and insertions), DocWhisper corrected substitution errors the most.
To delve into the details, we collected 100 instances that are substitution errors in Whisper but correct in DocWhisper and categorized them into four groups: technical term, inflection, mishearing, and name.
While the anticipated large proportion (41\%) of technical terms was observed, noteworthy percentages were also found for inflection (28\%) and mishearing (24\%). 
Many words with high familiarity could result in lower ranks when sorting based on word frequency, potentially causing a decline in the performance of FQ Ranker.
We show the error types and specific examples in Table~\ref{tab:error analysis} and more details in Appendix~\ref{appendix:error analysis}.
\section{Conclusion and Future Work}
We constructed an AVSR dataset, SlideAVSR, by utilizing paper explanation videos.
We proposed DocWhisper, which leverages OCR to refer to slide content.
We verified the effectiveness of DocWhisper on SlideAVSR and conducted a detailed analysis.
Additionally, we introduced FQ Ranker, which calculates word ranks based on word frequency, and evaluated its performance on DocWhisper.

In the future, we plan to continually refine OCR-based methods and aim to construct an end-to-end AVSR model that is not dependent on OCR.
Furthermore, we intend to build a benchmark that allows a comprehensive evaluation of the image comprehension capabilities of AVSR models by incorporating diverse types of videos, such as sports commentary, gaming commentary, cooking videos, and more.
Ultimately, we aim to construct a foundation model for AVSR that exhibits high performance across diverse video inputs.
\section*{Limitations}
In comparison to mainstream AVSR datasets, SlideAVSR exhibits a notably limited number of videos and speakers.
This may lead to data imbalance and create obstacles to the model's training process.
In addition, due to our focused collection of scientific paper explanation videos related to artificial intelligence, imbalances may have emerged in terms of speaker nationality, age, and gender.

Compared to SlideSpeech~\cite{wang2023slidespeech}, we introduced a vision-language model to filter videos and utterances that do not correspond to the slides, thereby reducing annotation costs and improving the quality of the dataset.
However, our dataset construction process still relies on manual annotation. Fully automating this process will be a major challenge for the future.

In Section~\ref{sec:data partition}, we attempted to classify speakers' accents by collaborating with native English speakers.
However, the task of assigning precise labels to every video was impeded by the complexity of distinguishing certain speakers' accents.
As a result, we selectively picked out videos with South Asian English accents, leaving the remainder unlabeled.
Ideally, each data split should exhibit a comparable distribution of accents, but this was unattainable due to the aforementioned challenges.
\section*{Ethical Considerations}
In adherence to the terms of use and copyright policies governing the YouTube platform, we collected data exclusively from publicly available videos.
We acknowledge the potential presence of sensitive information in our dataset, such as personal names and portraits.
To prioritize privacy and responsible data sharing, we plan to release OCR results and public video URLs instead of raw video files.
Furthermore, the release of our dataset will be strictly limited to research purposes.
\bibliography{custom}

\clearpage
\onecolumn
\appendix
\section{The list of target conferences used in data collection}
\label{appendix:conf_list}
We show our target conferences in Table~\ref{tab:conf_list}.

\begin{table}[h]
\centering
\resizebox{0.4\columnwidth}{!}{%
\begin{tabular}{l|l}
\hline
Topic & Conference \\
\hline
NLP & ACL,\; NAACL,\; EMNLP \\
CV & CVPR,\; ICCV,\; ECCV \\
Speech & INTERSPEECH,\; ICASSP \\
AI & AAAI,\; IJCAI \\
ML & ICLR,\; ICML,\; NeurIPS \\
Data Mining & KDD,\; WSDM,\; WWW \\
Database & SIGMOD,\; VLDB,\; ICDE \\
IR & SIGIR \\
HCI & CHI \\
\hline
\end{tabular}}
\caption{Target conferences.}
\label{tab:conf_list}
\end{table}
\section{Models and prompts used in data filtering and cleansing}
\label{appendix:data_model_and_prompt}
We introduce the details of the models and prompts employed in the ChatGPT filter, BLIP-2 filter, and CTC alignment as described in Section~\ref{sec:filtering} and~\ref{sec:cleansing}.
\paragraph{ChatGPT filter.}
We used gpt-3.5-turbo.
The prompt we used is shown in Table~\ref{tab:chatgpt_prompt}.
\begin{table}[h]
\centering
\resizebox{0.75\columnwidth}{!}{%
\begin{tabular}{l}
\hline
Here is a description of a YouTube video:\\
\{DESCRIPTION\}\\
Using the description, check whether the video meets the following criteria.\\
- This video is a presentation video of a research paper.\\
- The description is written in English.\\
Attention, you can only answer 'Yes' or 'No' and you can only answer one time.\\
\hline
\end{tabular}}
\caption{Prompt for ChatGPT filter.}
\label{tab:chatgpt_prompt}
\end{table}
\paragraph{BLIP-2 filter.}
We used blip2-flan-t5-xl\footnote{https://huggingface.co/Salesforce/blip2-flan-t5-xl}.
The prompt we used is shown in Table~\ref{tab:blip_prompt}.
\begin{table}[h]
\centering
\resizebox{0.75\columnwidth}{!}{%
\begin{tabular}{l}
\hline
Question: This image is a screenshot of a video,\\
check whether the image meets the following criteria.\\
- It is a screen-sharing, not a photo shoot.\\
- It is a part of a slide for a research presentation.\\
Attention, you can only answer 'Yes' or 'No' and you can only answer one time.\\
Answer:\\
\hline
\end{tabular}}
\caption{Prompt for BLIP-2 filter.}
\label{tab:blip_prompt}
\end{table}
\paragraph{CTC alignment.}
We used kamo-naoyuki\_wsj\footnote{https://huggingface.co/espnet/kamo-naoyuki\_wsj} and ESPnet implemenations\footnote{https://github.com/espnet/espnet}.
\clearpage
\section{Implement details}
\label{appendix:implement details}
We fine-tuned both Whisper and DocWhisper using AdamW~\citep{loshchilov2019decoupled} with a learning rate of 2e-5, and we linearly warmed up the learning rate over 1,000 steps.
The batch size was set to 16.
Training was conducted for 10 epochs, and the checkpoint with the best performance on the Dev set was used for evaluation.
Additionally, training was performed with three different seed values, and the average was computed.
We performed text normalization\footnote{https://github.com/openai/whisper} for evaluation.
All experiments were conducted on a single NVIDIA A100 (40G) GPU.
\section{Specific examples}
\label{appendix:error analysis}
The corresponding screenshots to Table~\ref{tab:error analysis} are shown below, and the parts referred to in the correction are circled in red.

All the variations from the same lexical element, such as plural nouns, conjugated verbs, and third-person singular verbs, were classified as inflection.
If the label and prediction are not from the same lexical element, we classified the error as technical terms, mishearing, and names, respectively.

\begin{figure}[h]
    \centering
    \fbox{\includegraphics[width=0.7\linewidth]{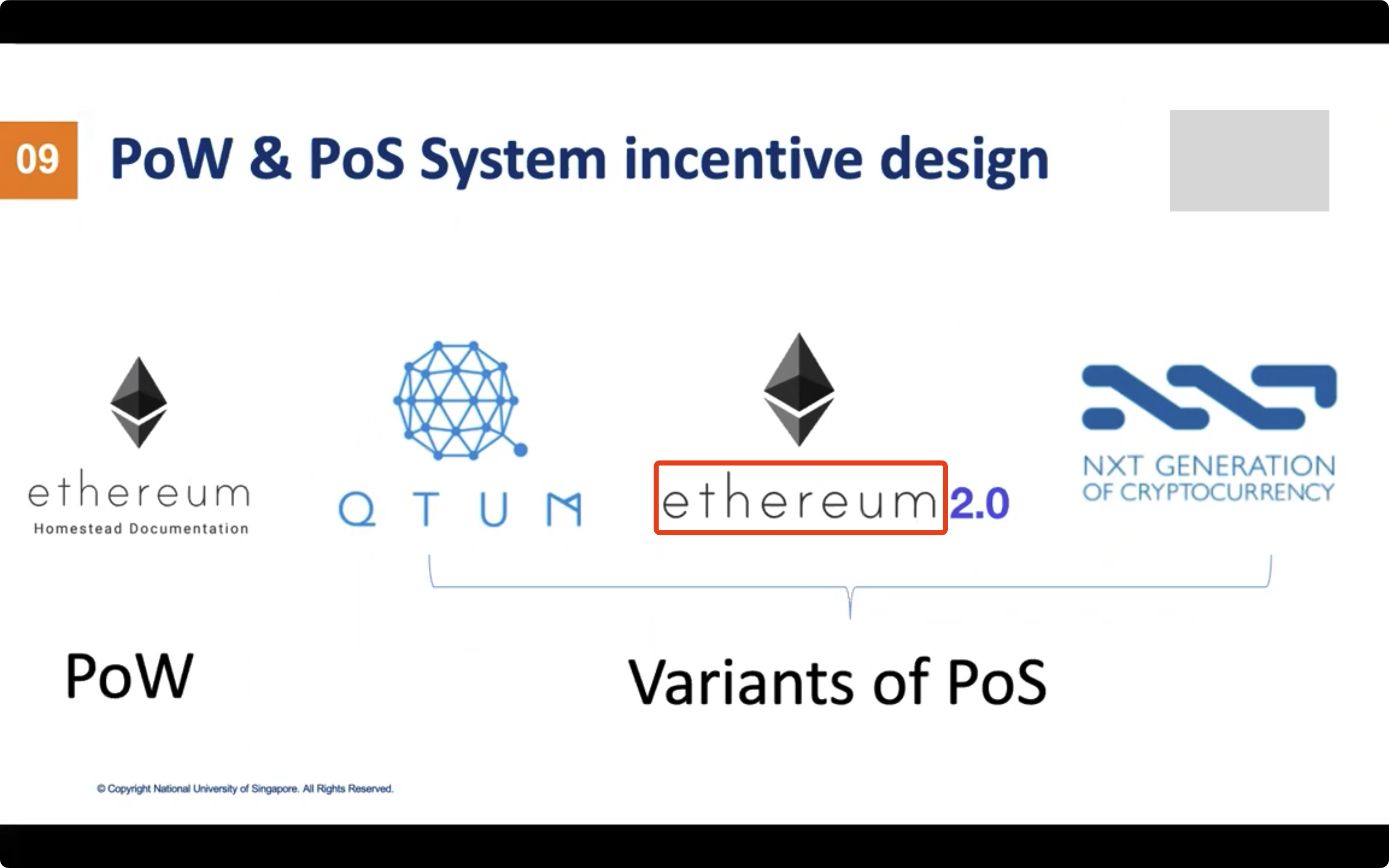}}
    \url{https://www.youtube.com/watch?v=eepUV9NJxFs}
\end{figure}
\begin{figure}[h]
    \centering
    \fbox{\includegraphics[width=0.7\linewidth]{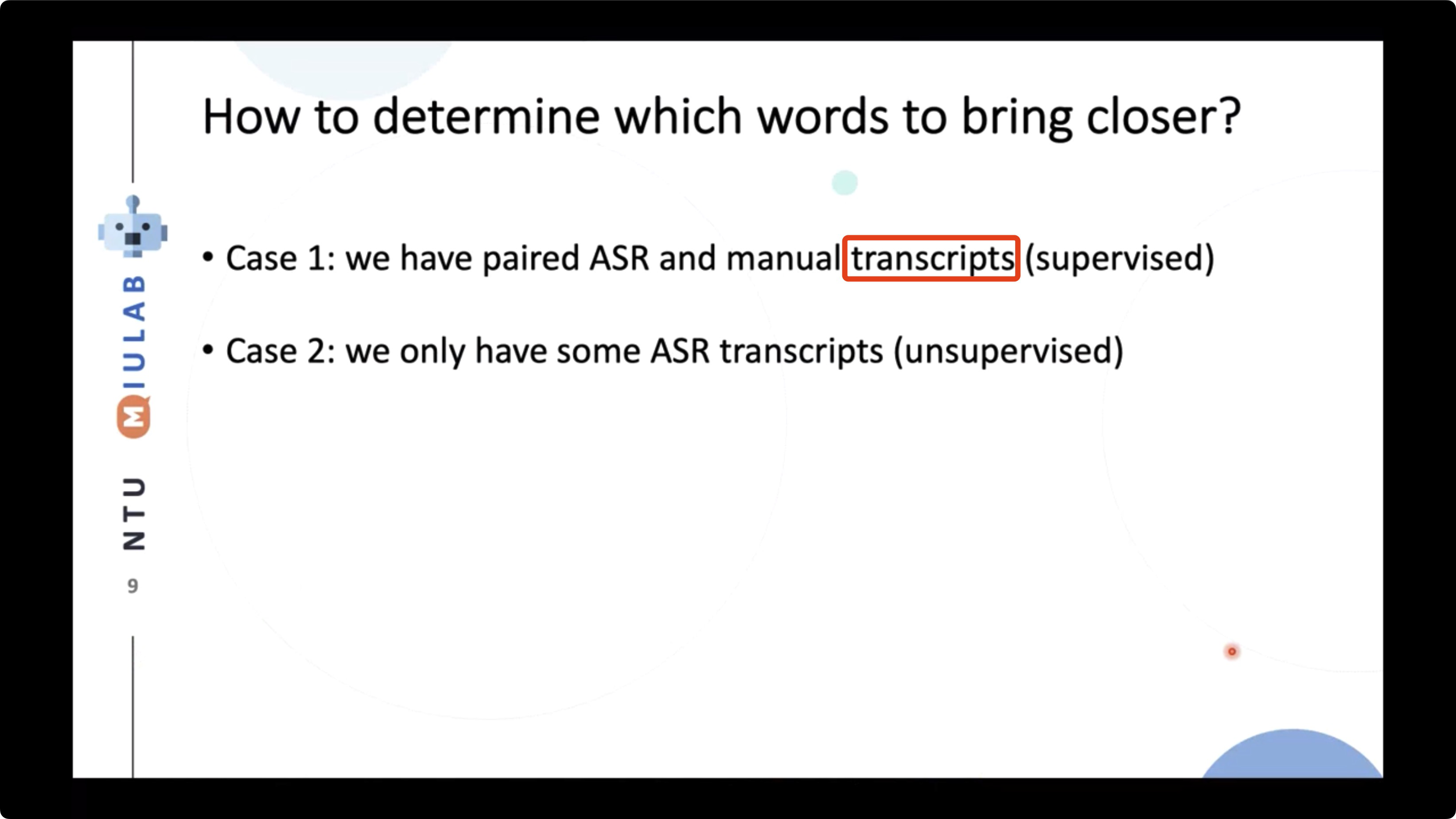}}
    \url{https://www.youtube.com/watch?v=dvUutyo72R4}
\end{figure}
\begin{figure}[t]
    \centering
    \fbox{\includegraphics[width=0.7\linewidth]{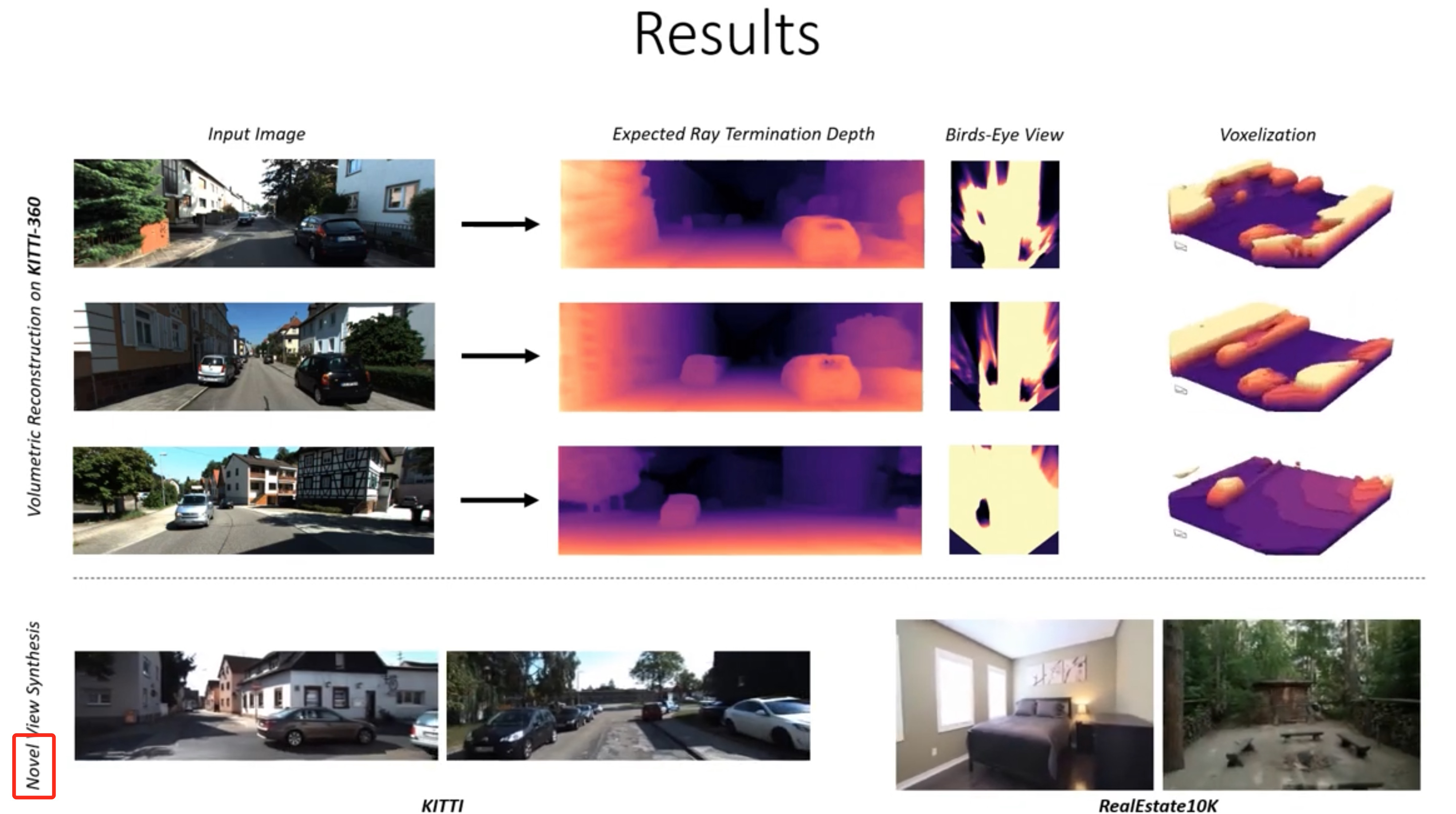}}
    \url{https://www.youtube.com/watch?v=0VGKPmomrR8}
\end{figure}
\begin{figure}[t]
    \centering
    \fbox{\includegraphics[width=0.7\linewidth]{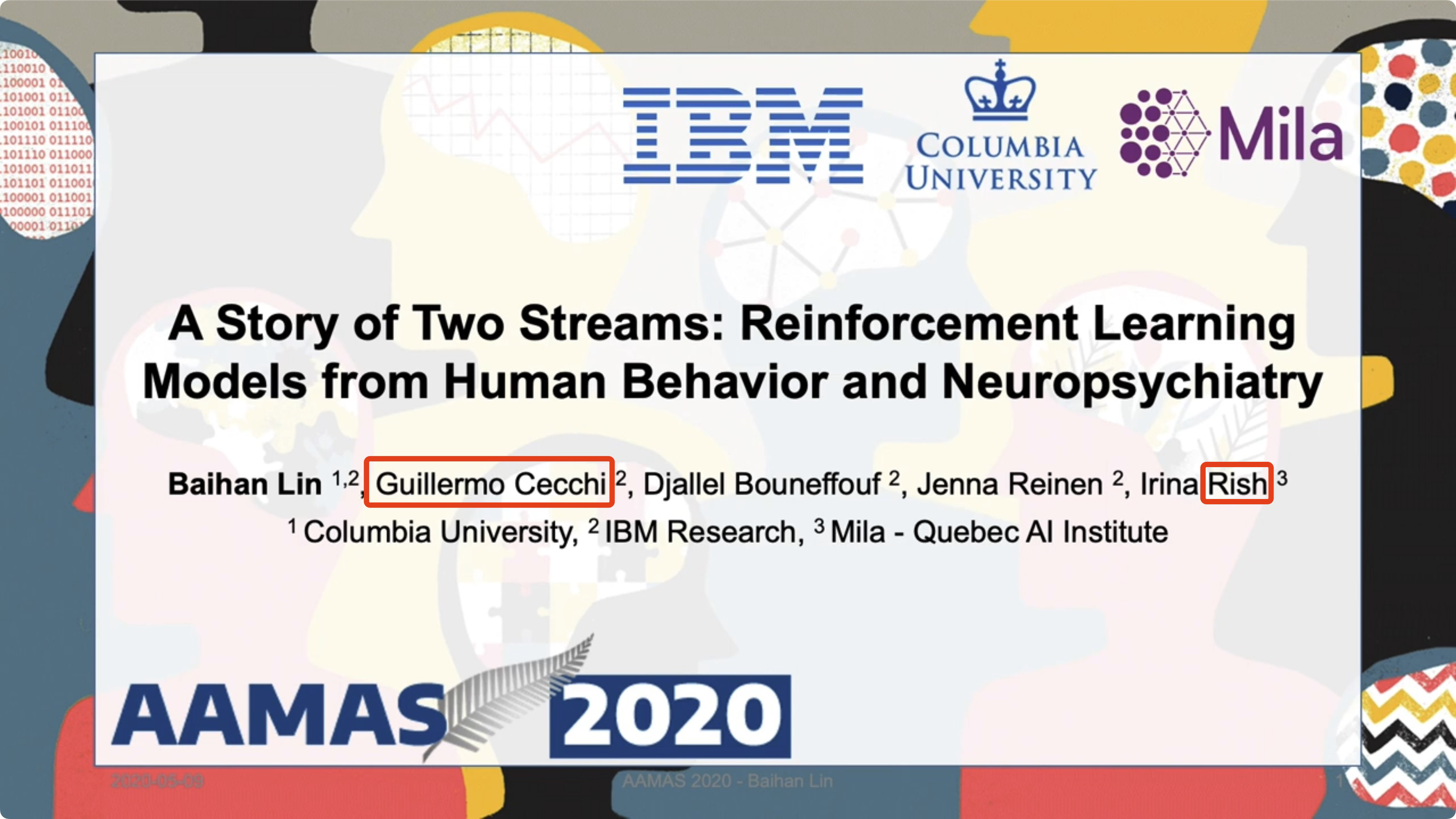}}
    \url{https://www.youtube.com/watch?v=CQBdQz1bmls}
\end{figure}
\end{document}